\documentclass[sigconf]{acmart}

\usepackage{booktabs}
\usepackage[acronym]{glossaries}
\usepackage{multirow}
\usepackage{subcaption}

\graphicspath{{Images/}}

\copyrightyear{2022}
\acmYear{2022}
\setcopyright{rightsretained}
\acmConference[ICONS 2022]{International Conference on Neuromorphic Systems}{July 27--29, 2022}{Knoxville, TN, USA}
\acmBooktitle{International Conference on Neuromorphic Systems (ICONS 2022), July 27--29, 2022, Knoxville, TN, USA}
\acmDOI{10.1145/3546790.3546794}
\acmISBN{978-1-4503-9789-6/22/07}

\newacronym {aer}   {AER}   {address-event representation}
\newacronym {adex}  {AdEx}  {Adaptive Exponential Integrate-and-Fire}
\newacronym {ai}    {AI}    {artificial intelligence}
\newacronym {ann}   {ANN}   {artificial neural network}
\newacronym {ap}    {AP}    {action potential}
\newacronym {api}   {API}   {application programming interface}

\newacronym {cam}   {CAM}   {content-addressable memory}
\newacronym {cmos}  {CMOS}  {complementary metal–oxide–semiconductor}
\newacronym {cnn}   {CNN}   {convolutional neural network}
\newacronym {cps}   {CPS}   {cyber-physical system}
\newacronym {cpu}   {CPU}   {central processing unit}

\newacronym {dl}    {DL}    {deep learning}
\newacronym {dnn}   {DNN}   {deep neural network}
\newacronym {dpi}   {DPI}   {differential-pair integrator}
\newacronym {dstc}  {dSTC}  {Disynaptic Spatiotemporal Correlator}
\newacronym {dynap} {DYNAP} {Dynamic Neuromorphic Asynchronous Processor}

\newacronym {e-i}   {E--I}  {excitatory--inhibitory}
\newacronym {epsp}  {EPSP}  {excitatory postsynaptic potential}
\newacronym {exc}   {Exc.}  {excitatory}

\newacronym {fdhm}  {FDHM}  {full duration at half maximum/minimum}
\newacronym {fpga}  {FPGA}  {field-programmable gate array}

\newacronym {gals}  {GALS}  {globally asynchronous locally synchronous}
\newacronym {gpu}   {GPU}   {graphics processing unit}

\newacronym {hil}   {HIL}   {hardware-in-the-loop}

\newacronym {ict}   {ICT}   {information and communications technology}
\newacronym {inh}   {Inh.}  {inhibitory}
\newacronym {iot}   {IoT}   {Internet of things}
\newacronym {ipsp}  {IPSP}  {inhibitory postsynaptic potential}
\newacronym {ipi}   {IPI}   {interpulse interval}
\newacronym {isi}   {ISI}   {interspike interval}

\newacronym {ltd}   {LTD}   {long-term depression}
\newacronym {ltp}   {LTP}   {long-term potentiation}

\newacronym {mos}       {MOS}       {metal oxide semiconductor}
\newacronym {mosfet}    {MOSFET}    {metal–oxide–semiconductor field-effect transistor}

\newacronym {nce}   {NCE}   {neuromorphic computing and engineering}

\newacronym {pir}   {PIR}   {postinhibitory rebound}
\newacronym {psp}   {PSP}   {postsynaptic potential}
\newacronym {pvt}   {PVT}   {power, voltage, and temperature}

\newacronym {stc}   {STC}   {Spatiotemporal Correlator}
\newacronym {snn}   {SNN}   {spiking neural network}
\newacronym {sram}  {SRAM}  {static random-access memory}
\newacronym {stdp}  {STDP}  {spike-timing-dependent plasticity}

\newacronym {tde}   {TDE}   {time-difference encoder}

\newacronym {vlsi}  {VLSI}  {very-large-scale integration}

\begin{document}

\title
[%
    % Header title
    Pattern Recognition
    in Heterogeneous Mixed-Signal Neurons
]
{%
    % Full title
    %Spatiotemporal Spike-Pattern Selectivity\\
    %in Mixed-Signal Silicon Neurons\\
    %"Utilizing" Hardware Heterogeneity
    %
    %Spatiotemporal Spike-Pattern Recognition\\
    %in Mixed-Signal Silicon Neurons\\
    %with Heterogeneous Dynamic Synapses
    %
    Spatiotemporal Pattern Recognition
    in Single Mixed-Signal VLSI Neurons
    with Heterogeneous Dynamic Synapses
}

\author{Mattias Nilsson}
\orcid{0000-0003-1024-5821}
\affiliation{%
    \department{EISLAB}
    \institution{Luleå University of Technology}
    \city{Luleå}
    \country{Sweden}
    \postcode{971 87}
}
\email{mattias.1.nilsson@ltu.se}

\author{Foteini Liwicki}
\orcid{0000-0002-6756-0147}
\affiliation{%
    \department{EISLAB}
    \institution{Luleå University of Technology}
    \city{Luleå}
    \country{Sweden}
    \postcode{971 87}
}

\author{Fredrik Sandin}
\orcid{0000-0001-5662-825X}
\affiliation{%
    \department{EISLAB}
    \institution{Luleå University of Technology}
    \city{Luleå}
    \country{Sweden}
    \postcode{971 87}
}

\renewcommand{\shortauthors}{%
    Nilsson,
    Liwicki,
    and Sandin
}

\begin{abstract}
%
% Aim for ~150 words?
%
% 1. Area
%Current developments within deep learning are pushing the limits of power-hungry digital computers in an unsustainable manner, while, conversely, digitalization is calling for increasingly low-power computation for decentralized edge AI.
%
%One possible way forward is the use of mixed-signal neuromorphic processors, which offer an ultra-low-power, non--von Neumann computational substrate for pattern recognition and learning by using subthreshold analog electronics to emulate electrochemical dynamics of the brain.
%
Mixed-signal neuromorphic processors with brain-like organization and device physics offer an ultra-low-power alternative to the unsustainable developments of conventional deep learning and computing.
%
% 2. Problem
However, realizing the potential of such neuromorphic hardware requires efficient use of its heterogeneous, analog neurosynaptic circuitry with neurocomputational methods for sparse, spike-timing-based encoding and processing.
%
% 3. Solution
%Here, we investigate the use of excitatory--inhibitory disynaptic lateral connections as a resource-efficient alternative to dedicated delay mechanisms in the thalamocortically inspired  Spatiotemporal Correlator (STC) spiking neural network.
%
Here, we investigate the use of balanced excitatory--inhibitory disynaptic lateral connections as a resource-efficient mechanism for implementing a thalamocortically inspired Spatiotemporal Correlator (STC) neural network without using dedicated delay mechanisms.
%
% 4. Methods
We present hardware-in-the-loop experiments with a DYNAP-SE neuromorphic processor, in which receptive fields of heterogeneous coincidence-detection neurons in an STC network with four lateral afferent connections per column were mapped by random input-sampling.
%
% 5. Results
%We find that, when the disynaptic elements are randomly programmed, some of the neurons display distinct receptive fields.
%
Furthermore, we demonstrate how such a neuron was tuned to detect a particular spatiotemporal feature by discrete address-reprogramming of the analog synaptic circuits.
The energy dissipation of the disynaptic connections is one order of magnitude lower per lateral connection (0.65~nJ vs 9.6~nJ per spike) than in the former delay-based hardware implementation of the STC.
%
% 6. Take-away
%Thus, we show how the heterogeneous neurosynaptic circuits can be used for resource-efficient implementation of event-driven STC-network layers in a way that enables synapse-address reprogramming as a discrete mechanism for feature tuning in analog-based hardware.
\end{abstract}

\begin{CCSXML}
<ccs2012>
    <concept>
        <concept_id>10010583.10010786.10010792.10010798</concept_id>
        <concept_desc>Hardware~Neural systems</concept_desc>
        <concept_significance>500</concept_significance>
    </concept>
    <concept>
        <concept_id>10010147.10010257.10010293.10010294</concept_id>
        <concept_desc>Computing methodologies~Neural networks</concept_desc>
        <concept_significance>500</concept_significance>
    </concept>
</ccs2012>
\end{CCSXML}

\ccsdesc[500]{Hardware~Neural systems}
\ccsdesc[500]{Computing methodologies~Neural networks}

% Separate the keywords with commas.
\keywords{Neuromorphic,
Spatiotemporal pattern recognition,
Ultra-low-power,
Device mismatch,
Neural heterogeneity,
Temporal code,
Excitatory--inhibitory balance}

\maketitle

\section{Introduction}
\label{sec:introduction}

% Neuromorphic Engineering

Neuromorphic engineering \cite{mead2020how, indiveri2021introducing} deals with the creation and use of physical substrates for information processing and sensing that imitate processes and structures observed in the brain.
Following this approach, mixed-signal neuromorphic processors are \gls{vlsi} systems that emulate the biophysical dynamics of neurons and synapses \cite{chicca2014electronic} in event-driven \glspl{snn} \cite{maass1997networks} and have high potential for ultra-low-power pattern recognition and learning \cite{rajendran2019low-power, davies2021advancing, frenkel2021bottom}.
Such technology can offer a sustainable alternative to deep learning \cite{thompson2021deep, mehonic2022brain} and enable always-on \textit{edge intelligence} \cite{zhou2019edge, ye2021challenges} in sensing applications, such as wearable biomedical devices \cite{covi2021adaptive}.
However, such analog-based systems are subject to transistor parameter-variance---so-called “device mismatch” \cite{pelgrom1989matching, tuinhout2010parametric}---which gives rise to substantial static parameter variance within uniformly configured populations of neurons and synapses \cite{qiao2016scaling}, and which poses a challenge to configurability and computational precision.
Furthermore, the dynamic real-time operation of the neurosynaptic computational nodes of these processing systems needs to be employed in a way that realizes their potential for sparse, event-driven information processing \cite{indiveri2019importance} with spike-timing-based encoding \cite{thorpe2001spike, guo2021neural, davidson2021comparison, stockl2021optimized, goltz2021fast}---as opposed to purely rate-based encoding, the use of which rather makes \glspl{snn} more or less analogous to the artificial neural networks often used within the deep learning paradigm.
%
% Spatiotemporal pattern recognition
%In \cite{haessig2020touch}, a neuromorphic model for classification of spatiotemporal spike patterns in an unsupervised manner is presented.
%
%Neuronal--synaptic delays play a central role in this and other related works on spatiotemporal pattern recognition \cite{agmon-snir1993delay, sheik2012exploiting, nielsen2017compact, sandin2020delays}.
%
%Other examples of theoretical frameworks for neural processing that require variability in the processing elements include ensemble learning, reservoir computing, and liquid state machines. 
%
%A number of current neuromorphic hardware implementations of such frameworks for spatiotemporal pattern recognition problems rely on the variability caused by transistor device-mismatch \cite{sheik2012emergent, richter2015random, das2018unsupervised, bauer2019real-time, donati2019discrimination, nilsson2020integration}.

% Spatiotemporal Pattern Recognition

Learning and recognition of spatiotemporal patterns---in contrast to static, purely spatial patterns, or even sequences of such---in artificial systems is an engineering problem for which spike-based neuromorphic computing holds particularly high potential for efficient solutions in terms of energy and latency \cite{davies2021advancing, haessig2020touch}.
A central aspect of such pattern recognition with \glspl{snn} is the neural encoding of temporal relations, which may be performed in a number of different ways \cite{sheik2013spatiotemporal}.
One bio-inspired approach to temporal encoding in \glspl{snn} is the use of neural spike-propagation delays \cite{agmon-snir1993delay, thorpe2001spike}, which have been implemented in or for neuromorphic hardware in the form of axonal delays \cite{nielsen2017compact}, delay interneurons \cite{sheik2012exploiting}, and emulation of dendrites \cite{wang2012active, kaiser2021emulating, benjamin2021neurogrid}.
However, such implementations come with a substantial cost, as they require allocation of additional hardware resources beyond what is already used to implement the neurons and synapses of a given \gls{snn} and sometimes beyond what is readily available in general-purpose neuromorphic hardware.
Another relevant concept is that of the spiking \gls{tde} \cite{milde2018elementary, haessig2020touch}, which is a versatile neurocomputational primitive for temporal encoding, but which cannot be implemented in general-purpose analog-based neuromorphic systems without dedicated hardware.

% Computing with Mismatch

The challenge to using mixed-signal neuromorphic hardware posed by the imprecision stemming from device mismatch has previously been addressed with usage of relatively costly compensation methods, such as circuit calibration procedures \cite{lenero2008calibration, neftci2010device} or neural resource redundancy \cite{kauderer2017population}.
Other mitigation methods include the use of neural--architectural robustness to noise and variability \cite{buchel2021supervised} and special training procedures that promote robustness during inference \cite{buchel2022network}.
In contrast, however, some current approaches to spatiotemporal pattern recognition with mixed-signal neuromorphic hardware are---rather than attempting to mitigate the variance caused by device mismatch---based on neural processing frameworks that actually rely on variability in the neurosynaptic elements \cite{sheik2012emergent, richter2015device, donati2018processing, das2018unsupervised, bauer2019real-time, nilsson2020integration}, such as reservoir computing and ensemble learning.
Similarly, Markov chain Monte Carlo sampling of heterogeneous memristors has been used to train a Bayesian machine learning model \cite{dalgaty2021situ}.
Furthermore, there is recent research suggesting that the existence of neurosynaptic variability, which is also observed in the brain, may support efficiency and robustness in neural processing and learning---especially for information that has a rich temporal structure \cite{perez2021neural, zeldenrust2021efficient}.

% Receptive-Fields Experiment

%Here, we investigate event-driven, spike-timing-based spatiotemporal pattern recognition in ultra-low-power, albeit heterogeneous, mixed-signal neuromorphic hardware.
%
Here, we follow the approach of making use of device mismatch as a source of variance for resource-efficient employment of inherently inhomogeneous mixed-signal neuromorphic hardware.
We investigate usage of a previously proposed form of \gls{e-i} balanced disynaptic elements \cite{sandin2020delays} as a heterogeneity-dependent low-resource alternative to dedicated delay-mechanisms for coincidence-detection-based spatiotemporal pattern recognition with a modified version of the \gls{stc} network, which we propose here---the \gls{dstc}.
%
%We selected the \gls{stc} neural network as a starting point for our investigation based on its relative simplicity and intelligibility, compatibility with general-purpose neuromorphic hardware, and potential to form layers in deep \glspl{snn} for processing of increasingly complex spatiotemporal patterns.
%
To investigate the feasibility and effectiveness of implementing the \gls{dstc} in mixed-signal neuromorphic hardware, we present hardware-in-the-loop experiments with a DYNAP-SE neuromorphic processor \cite{moradi2018dynaps}, in which we characterized spike-timing-based spatiotemporal receptive fields that form in coincidence-detection neurons of the \gls{dstc}.
Furthermore, we performed tuning of one of these receptive fields by using digital synapse-address reprogramming to sample from the inherent parameter distributions of the analog synapses.
Finally, we present an estimate of the difference in energy usage for hardware implementations of the \gls{dstc} and \gls{stc} neural networks, which indicates a reduction by one order of magnitude per lateral connection (0.65~nJ vs 9.6~nJ per spike).

% Conclusion

In summary, we present how heterogeneous neurosynaptic dynamics in neuromorphic processors like the DYNAP-SE can be used in a resource-efficient manner for spike-timing-based spatiotemporal pattern recognition in a way that enables synapse-address reprogramming as a discrete mechanism for feature tuning, which results in observable and reproducible state changes.
This approach may serve as a complement to more accurate but resource-intensive delay-based coincidence detection or dendritic integration and use of volatile plastic synapses, which hamper observability and reproducibility.
Our study also contributes a new perspective on the \gls{stc}'s mechanism and capability for pattern recognition on the level of single neurons---as opposed to the population-level view held in the original studies \cite{sheik2012emergent, coath2014robust}.

\section{Materials and Methods}
\label{sec:methods}
% Experimental setup
The experiments presented in this work were conducted with a DYNAP-SE (Dynamic Neuromorphic Asynchronous Processor -- Scalable) \cite{moradi2018dynaps} from SynSense interfaced in a closed loop with a PC using the software Legacy~Samna \cite{samna} (formerly CTXCTL).
All input stimuli were generated using the built-in spike-generator in the \gls{fpga} of the DYNAP-SE, which generates spike-events according to assigned temporal \glspl{isi} and virtual source-neuron addresses.

% Footnotes
%\addtocounter{footnote}{-1}
%\footnotetext{https://www.synsense-neuromorphic.com/technology}
%\stepcounter{footnote}
%\footnotetext{https://pypi.org/project/samna/}

\subsection{Neuromorphic Processor}

\begin{figure}[tb]
    \centering
    \includegraphics[width=\columnwidth]{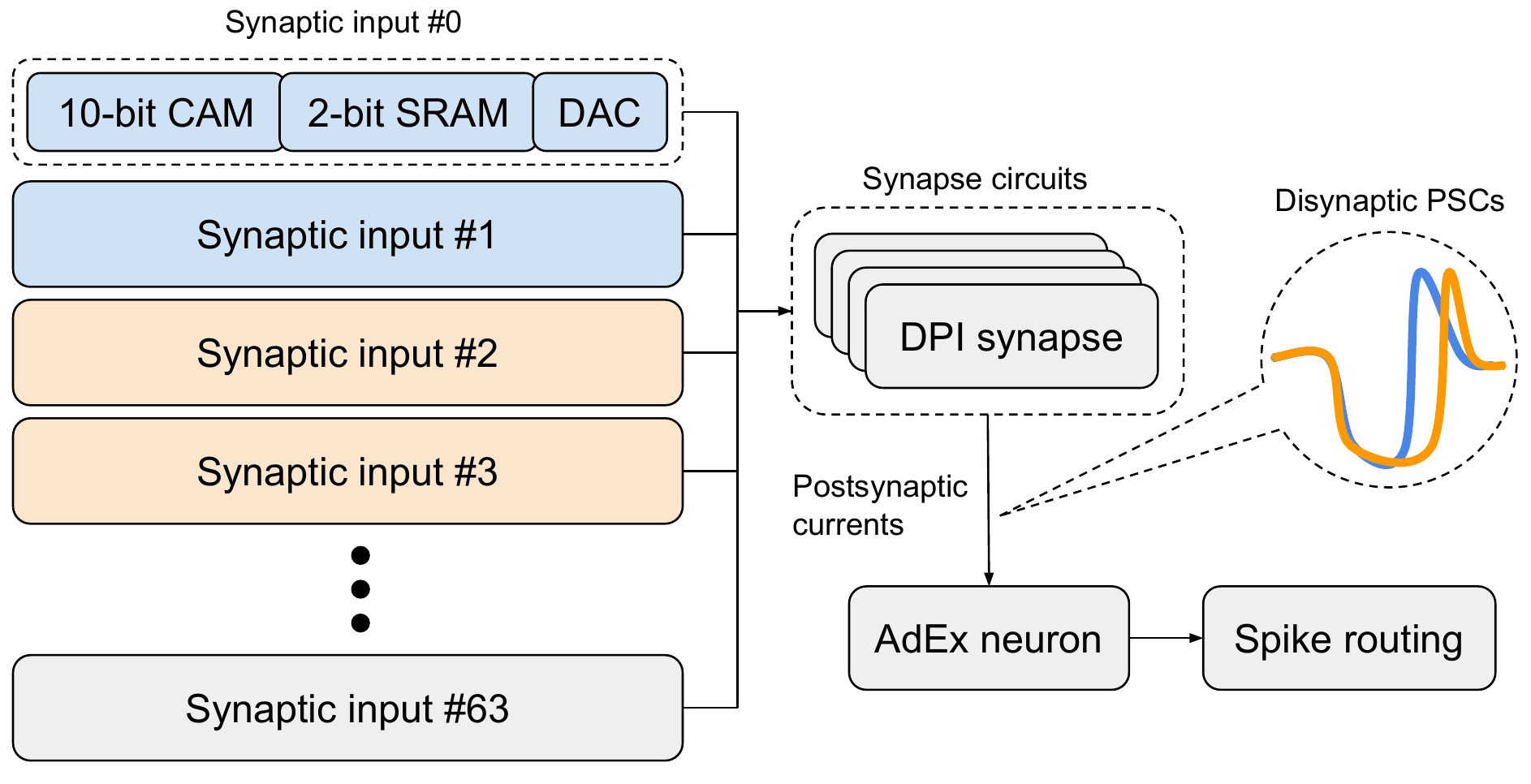}
    \caption{%
        \textbf{%
            Implementation of balanced excitatory--inhibitory (E--I) disynaptic elements in one neuron of the DYNAP-SE neuromorphic processor.
        }%
        Based on a figure from \cite{moradi2018dynaps}.
        Each neurosynaptic node contains 64 synaptic input-channels---each with a CAM cell for the presynaptic neuron address, an SRAM cell for the synapse-type information, and digital-to-analog signal-converting circuitry (DAC)---four analog synaptic DPI circuits of different types, and one analog AdEx-neuron circuit.
        The coloring indicates the use of two synaptic inputs for each E--I element \cite{sandin2020delays, nilsson2020integration}.
        Two examples of the resulting heterogeneous postsynaptic currents (PSCs) for different E--I elements are illustrated, which, respectively, use synaptic inputs 0--1 and 2--3.
        The E--I elements are used for synapse-based generation of delayed excitations for coincidence detection, instead of using dedicated delay lines or emulation of dendrites.
        %
        %\todo{In Figs.~3--5 input channel indices run from $0$ and up, and is defined differently than here. Perhaps the dashed box can enclose both blue boxes, and the label can be changed to ``input channel \#0''. The text inside the boxes could be changed to ``Synapse \#2'' etc, to distinguish synapse indices from input channel indices.
        %
        %Could we somehow illustrate that the pairs of colored synapse rows are E--I pairs? For example by keeping only two DPI blocks, one labeled E and one I (``DPI E-synapse'' or similar), and making two arrow-lanes from the colored boxes to each of the DPI blocks?
        %
        %Briefly define also the abbreviations DACC, SRAM, DPI in Sect. II.A, and explain that the difference between blue/red delayed excitations appears due to device mismatch in the DACC blocks.
        %For clairty, make all synapse boxes equally wide as the first row.} 
    }
    \label{fig:dynap_node}
\end{figure}

% DYNAP-SE overview
The DYNAP-SE \cite{moradi2018dynaps} is a reconfigurable, general-purpose, mixed-signal neuromorphic processor that uses subthreshold analog circuits to emulate the biophysical dynamics of neurons and synapses in real time, and asynchronous digital circuits for spike-event transmission according to an \gls{aer} protocol.
One DYNAP-SE unit comprises four four-core neuromorphic chips---each of which comprises 256 silicon neuron-circuits implementing the \gls{adex} spiking neuron model \cite{brette2005adex}.
Each neuron has a \gls{cam} block that can contain up to 64 different addresses, each representing a presynaptic neuronal connection\footnotemark, see \textbf{Figure~\ref{fig:dynap_node}}.
For each connection, dynamic synapses in the form of \gls{dpi} circuits \cite{bartolozzi2007synaptic} are available in four different synaptic types: fast and slow excitatory, and subtractive and shunting inhibitory.
The dynamic behaviors of the neuronal and synaptic circuits of the DYNAP-SE are governed by analog circuit parameters that are set by on-chip programmable bias generators, which provide 25 bias parameters independently for each core.
In the present work, a certain kind of balanced \gls{e-i} disynaptic elements \cite{sandin2020delays, nilsson2020integration} are used to generate delayed excitations in the DYNAP-SE, in a way that is further described in Section~\ref{sec:dSTC}.

\footnotetext
{%
    Each CAM address can match the same local neuron-ID on different cores, which enables multiple presynaptic connections per input circuit.
}

\subsection{Spiking Neural Network}

% STC Intro 
%Biological neural systems can, in addition to informing neuromorphic hardware design, also provide inspiration for architectural principles of neural network models for efficient processing \cite{dalgaty2021bio-inspired}.
%
The \acrfull{stc} is a \acrfull{snn} for spatiotemporal pattern recognition and learning in neuromorphic hardware---based on coincidence detection of lateral, temporally delayed projections \cite{sheik2012exploiting}---which was derived from a biologically plausible model of thalamocortical auditory processing \cite{coath2011thalamocortical} and has been implemented in real-time mixed-signal neuromorphic hardware \cite{sheik2012emergent, sheik2013thesis}.
The originators of the \gls{stc} argue that the functional principles of the network and of its more complex precursor can be used to produce spike-based feature extractors, which may form the basis of sensory--processing systems based on mixed-signal neuromorphic technology.
Furthermore, the original conception and study of the \gls{stc} was also motivated as a case-study of the neural mechanisms underlying feature selectivity in primary sensory processing---especially in the auditory domain---as well as a study of the intrinsic characteristics of such features.
The robustness of the \gls{stc} to variability in stimulus patterns---which is a prerequisite for most real-world sensing applications---has been further demonstrated in a subsequent study \cite{coath2014robust}.

\begin{figure*}[tb]
    \centering
    \begin{subfigure}[b]{0.9\columnwidth}
        \includegraphics[width=\textwidth]{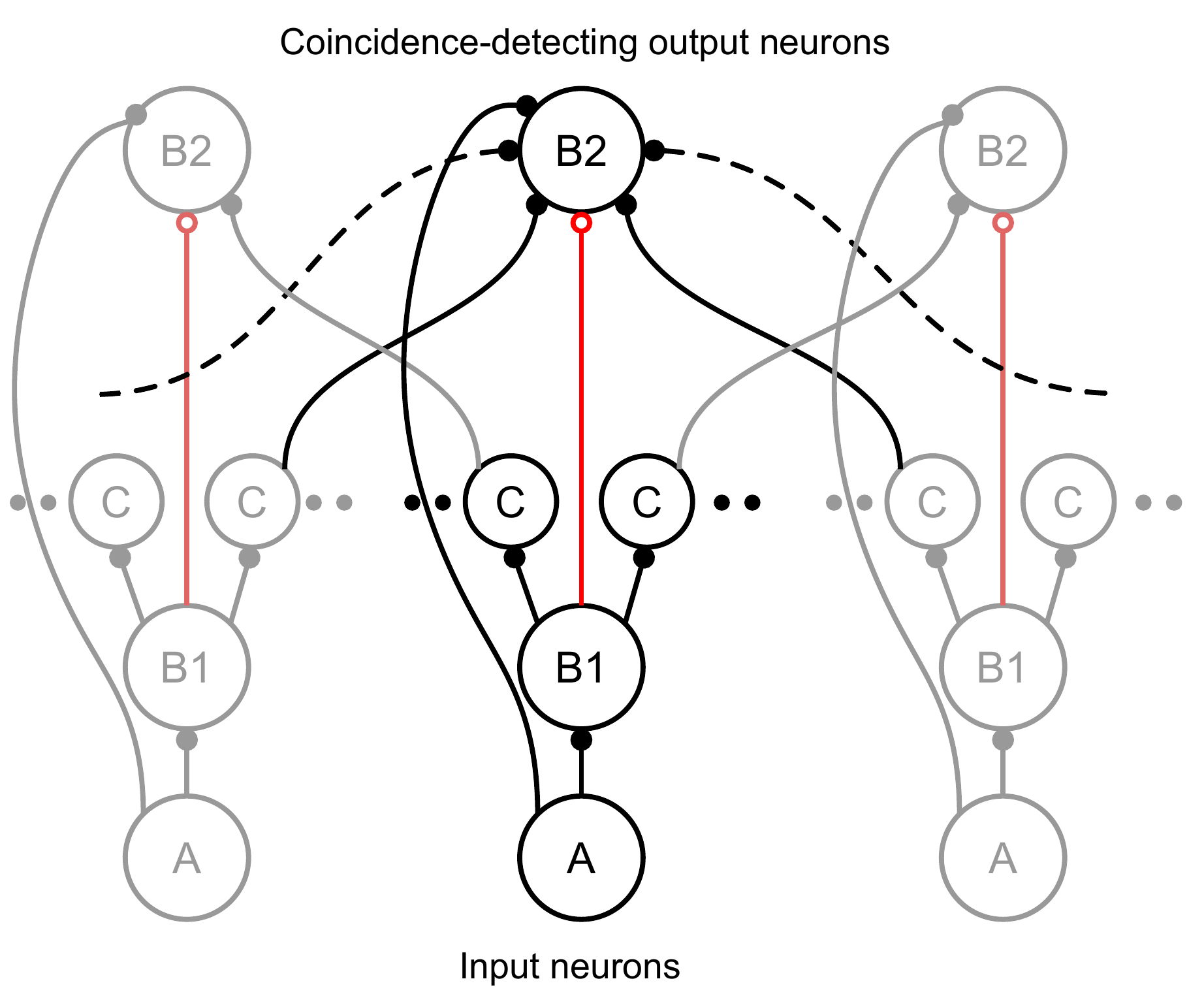}
        %\put(-200, 180){\textbf{A}}
        \caption{}
        \label{fig:STC}
    \end{subfigure}
    \hfill
    \begin{subfigure}[b]{1.05\columnwidth}
        %\hspace{1.2cm}
        \includegraphics[width=\textwidth]{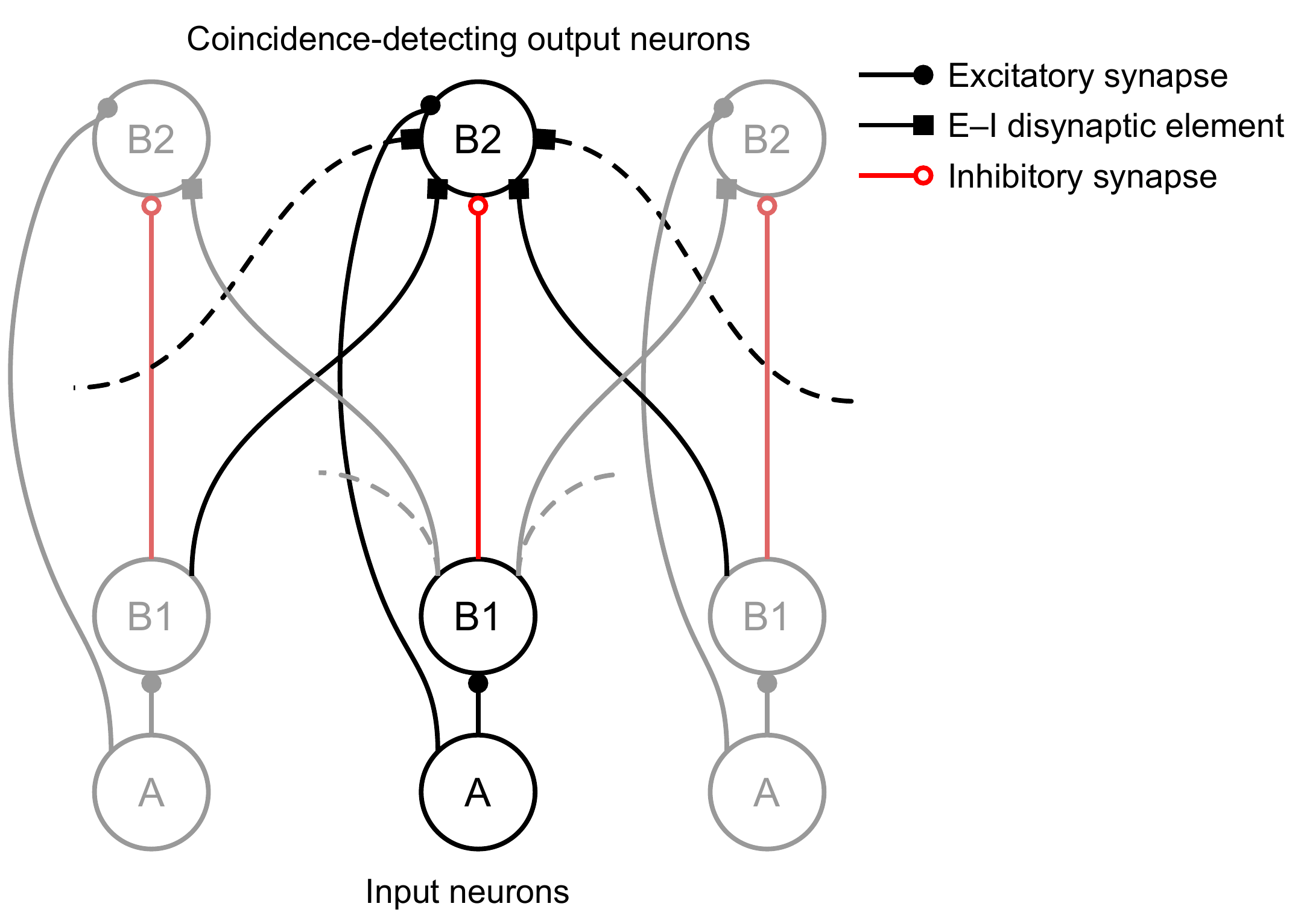}
        %\put(-225, 180){\textbf{B}}
        \caption{}
        \label{fig:sSTC}
    \end{subfigure}
    \caption{%
        \textbf{%
            Spatiotemporal Correlator (STC) spiking neural network--architectures.
        }%
        \textbf{(a)}
        Original STC network, adapted from \cite{sheik2013thesis}.
        \textbf{(b)}
        Disynaptic STC (dSTC) network, which uses excitatory--inhibitory (E--I) disynaptic elements \cite{sandin2020delays, nilsson2020integration}.
        For each architecture, one column is highlighted, and the two closest neighboring columns are included in shaded colors to illustrate lateral connections.
        Circles represent single neurons from populations A, B1, B2, and C.
    }
    \label{fig:STC_nets}
\end{figure*}

% STC Topology and Mechanism
%The Spatiotemporal Correlator (STC) neural network---originally proposed in \cite{sheik2012emergent} and further described in \cite{sheik2013thesis}---is a spiking neural network (SNN) architecture for spatiotemporal pattern learning and recognition that is based on the mechanism of coincidence detection of lateral, temporally delayed neuronal projections.
%
\vbox{%
    The \gls{stc} network, see \textbf{Figure~\ref{fig:STC}}, consists of the following qualitatively distinct neuronal populations:
    \begin{description}
        \item[\textbf{A:}] Input neurons
        \item[\textbf{B1:}] Secondary input neurons
        \item[\textbf{B2:}] Coincidence-detection neurons
        \item[\textbf{C:}] Delay neurons
    \end{description}
}
The network is structured in columns---not to be confused with cortical columns---each consisting of one A, B1, and B2 neuron, respectively.
Within each column, the input neuron, A, receives a signal from an input channel---such as a pixel of a visual sensor, or a frequency band of an auditory sensor---which it then projects to both the B1 and B2 neurons via excitatory synapses.
B1 then generates a one-to-one spike-response to its input from A, which is projected by inhibition to the B2 neuron of the same column and by excitation to the B2 neurons of some number of adjacent columns---the number of which is subject to design-choice and pruning.
Each lateral B1--B2 excitation is projected via a neuron from the C population, thereby inducing a temporal signal-propagation delay.
The A--B2 excitation is slightly faster than the B1--B2 inhibition, thus creating a time-window during which B2 is primed to spike in response to coincident lateral projections from adjacent columns.
Hence, each B2 neuron constitutes a coincidence detector that is sensitive to some particular set of spatiotemporal spike-patterns in a local sensory region, which forms a feature of the pattern that is recognized by the \gls{stc} network as a whole.

% https://docs.google.com/drawings/d/1zt5AuqdsIRveIVQ9zYNVHFfBTGOB9Y132tXy2NVu02Y/edit
%

\subsubsection{The Disynaptic STC Network}
\label{sec:dSTC}

Here, we propose a modified version of the \gls{stc}---the Disynaptic Spatiotemporal Correlator (dSTC), see \textbf{Figure~\ref{fig:sSTC}}.
In the \gls{dstc}, the delay interneurons of the \gls{stc} are replaced with balanced \gls{e-i} disynaptic elements---first proposed in \cite{sandin2020delays} and further investigated in \cite{nilsson2020integration}---thereby theoretically reducing the amount of resources required to implement the network by a substantial amount.
Each \gls{e-i} element consists of one excitatory and one inhibitory synapse connected to the same presynaptic neuron.
Thus, Dale's law is explicitly broken to save resources.
The time-constants and weights of the two synapses are balanced in such a way that each presynaptic spike generates---by summation of the postsynaptic currents---an inhibition followed by a postinhibitory excitation, thus generating a delayed excitation that imitates postinhibitory rebound \cite{sandin2020delays}.
When implemented on different neuronal and synaptic hardware circuits in a mixed-signal system such as the DYNAP-SE, see \textbf{Figure~\ref{fig:dynap_node}}, the \gls{e-i} elements generate varying temporal delays following a roughly Gaussian distribution, as demonstrated in \cite{sandin2020delays} and  \cite{nilsson2020integration}.
This variation is the source of the differences in temporal delay between the lateral afferent connections of B2 neurons in the \gls{dstc} architecture.
These differences make each B2 neuron sensitive to some range of spatiotemporal spike-patterns that are illustrated as receptive fields in the following, and which are dependent on the heterogeneity of the analog hardware circuits.
In this manner, the \gls{dstc} performs coincidence-based spatiotemporal feature detection by relying on synaptic integration of the type investigated in \cite{nilsson2020integration}.

\subsection{Mapping of Receptive Fields}
\label{sec:RF_mapping}

\begin{figure}[tb]
    \centering
    \begin{subfigure}[b]{0.45\columnwidth}
         \centering
         \includegraphics[width=\textwidth]{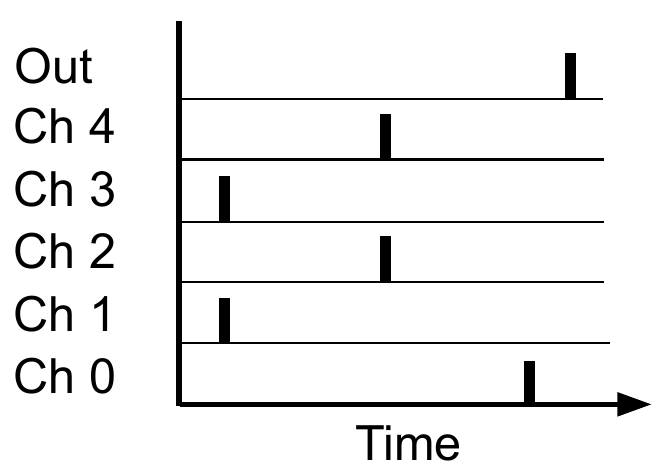}
         \caption{}
         \label{fig:rf_spikes}
     \end{subfigure}
     \hfill
     \begin{subfigure}[b]{0.45\columnwidth}
         \centering
         \includegraphics[width=\textwidth]{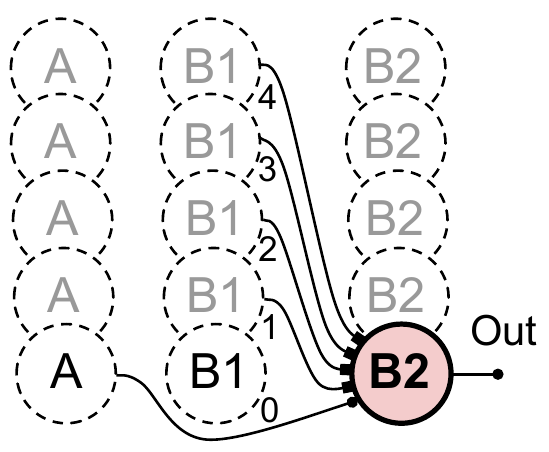}
         \caption{}
         \label{fig:rf_network}
     \end{subfigure}
    \caption{%
        \textbf{%
            Receptive-field experiment.
        }%
        \textbf{(a)}
        Example input spike-pattern and response.
        \textbf{(b)}
        Feature-detection neuron of dSTC network.
        The B2 neurons receive spatiotemporal spike patterns on five input channels, of which Channels 1--4 are lateral connections with, in effect, delayed excitations via E--I disynaptic elements.
    }
    \label{fig:rf_experiment}
\end{figure}

% Experiment outline
We investigated spike-based spatiotemporal receptive fields of the kind described in the previous section in B2 neurons, see \textbf{Figure~\ref{fig:rf_network}}, that have four lateral connections each---similar to the original \gls{stc} after training \cite{sheik2012emergent}---by implementing a population of B2 neurons in one core of the DYNAP-SE neuromorphic processor.
We focused on the forward and lateral connections and, for simplicity, omitted the B1--B2 inhibition, since its function is primarily to regulate spike-timing-dependent plasticity and to prevent sensitivity to excessive stimulation, and is therefore not required for an initial investigation of the receptive fields.

% Source: https://docs.google.com/drawings/d/1-inc_GHVUmbjj_eIDoKcP_nUET_XeCjJCcsjzy5Cytg/edit

% Experiment description
The receptive fields were mapped by stimulating the B2 neurons---independently from each other---with randomized spike-patterns ($N$~=~10,000), each consisting of one spike per input channel, as illustrated by the sample pattern in \textbf{Figure~\ref{fig:rf_spikes}}.
The spike-times of each of the lateral, delayed projections were independently drawn from a uniform random distribution ranging from 1--50~ms before the incidence of the direct forward excitation from A to B2, which we let define the reference time, $t$~=~0, of the pattern.
All spike-patterns for which a given B2 neuron generated one or more postsynaptic spikes in response were recorded and aggregated to form an approximation of the receptive field of that neuron.

\subsection{Feature Tuning by Synapse Sampling}
\label{sec:methods:tuning}

\begin{figure}[tb]
    \centering
    \includegraphics[width=\columnwidth]{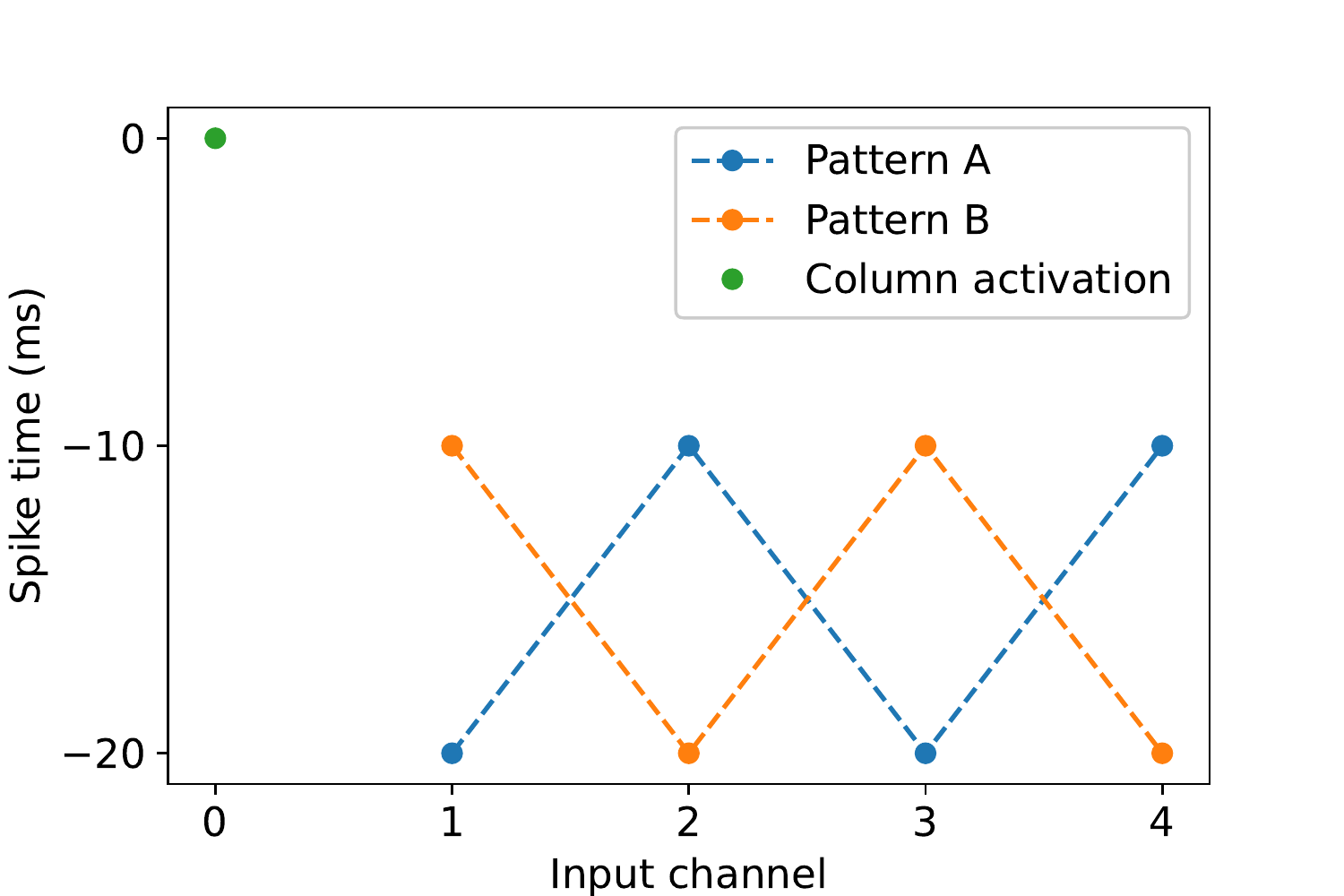}
    \caption{%
        \textbf{%
            Spatiotemporal spike patterns for feature tuning.
        }%
        Each pattern consists of one spike per input channel, including the feed-forward input at reference time $t$ = 0, resulting in five spikes per pattern.
    }
    \label{fig:tuning_patterns}
\end{figure}

Based on the results in \cite{nilsson2020integration}, we also investigated the possibility of tuning one of the receptive fields from the experiment described in Section~\ref{sec:RF_mapping} by changing the hardware configuration of the B2 neuron in question.
More specifically, we investigated whether replacing the specific heterogeneous hardware circuits used for each synaptic input-connection of the neuron, see \textbf{Figure~\ref{fig:dynap_node}}, could affect its coincidence detection enough to make the neuron distinguish between two different prescribed patterns, see \textbf{Figure~\ref{fig:tuning_patterns}}, which the neuron initially could not.
The synaptic reconfiguration was made by assigning to each of the synapses a random, unique input circuit drawn from the set of all the 64 input circuits of the neuron.
Following each subsequent synaptic reconfiguration, the neuron was presented with both of the different patterns, separately, ten times---and its response in terms of the number of postsynaptic spikes was recorded.

\section{Results}
\label{sec:results}
\subsection{Receptive Fields}

\begin{figure}[tb]
    \centering
    %
    %\vspace{-0.3cm}
    \begin{subfigure}[b]{0.49\columnwidth}
        \includegraphics[width=\textwidth]{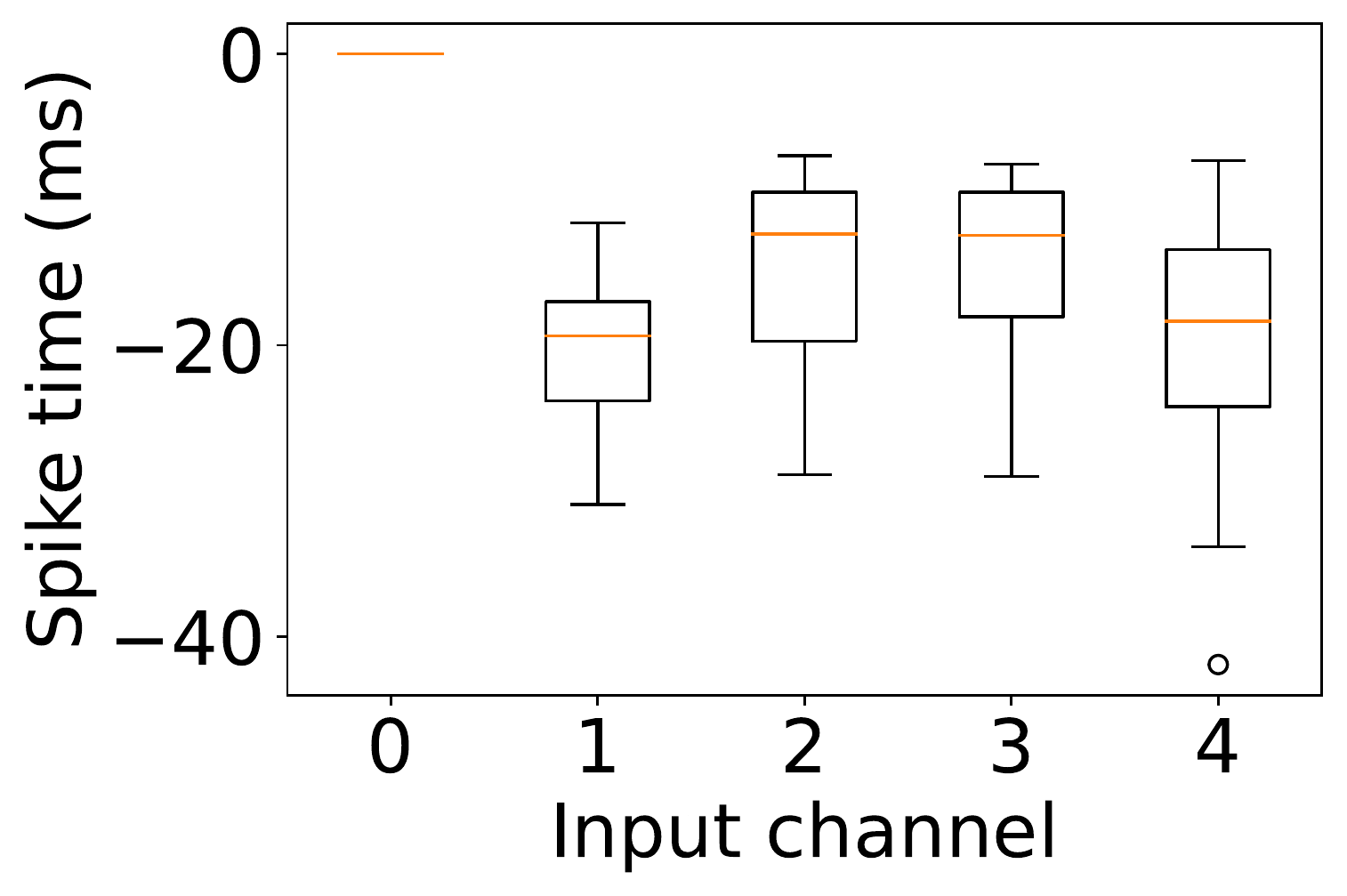}
        %\put(-95, 85){\textbf{A}}
        \caption{}
        \label{fig:rf_1}
    \end{subfigure}
    %\hfill
    \begin{subfigure}[b]{0.49\columnwidth}
        \includegraphics[width=\textwidth]{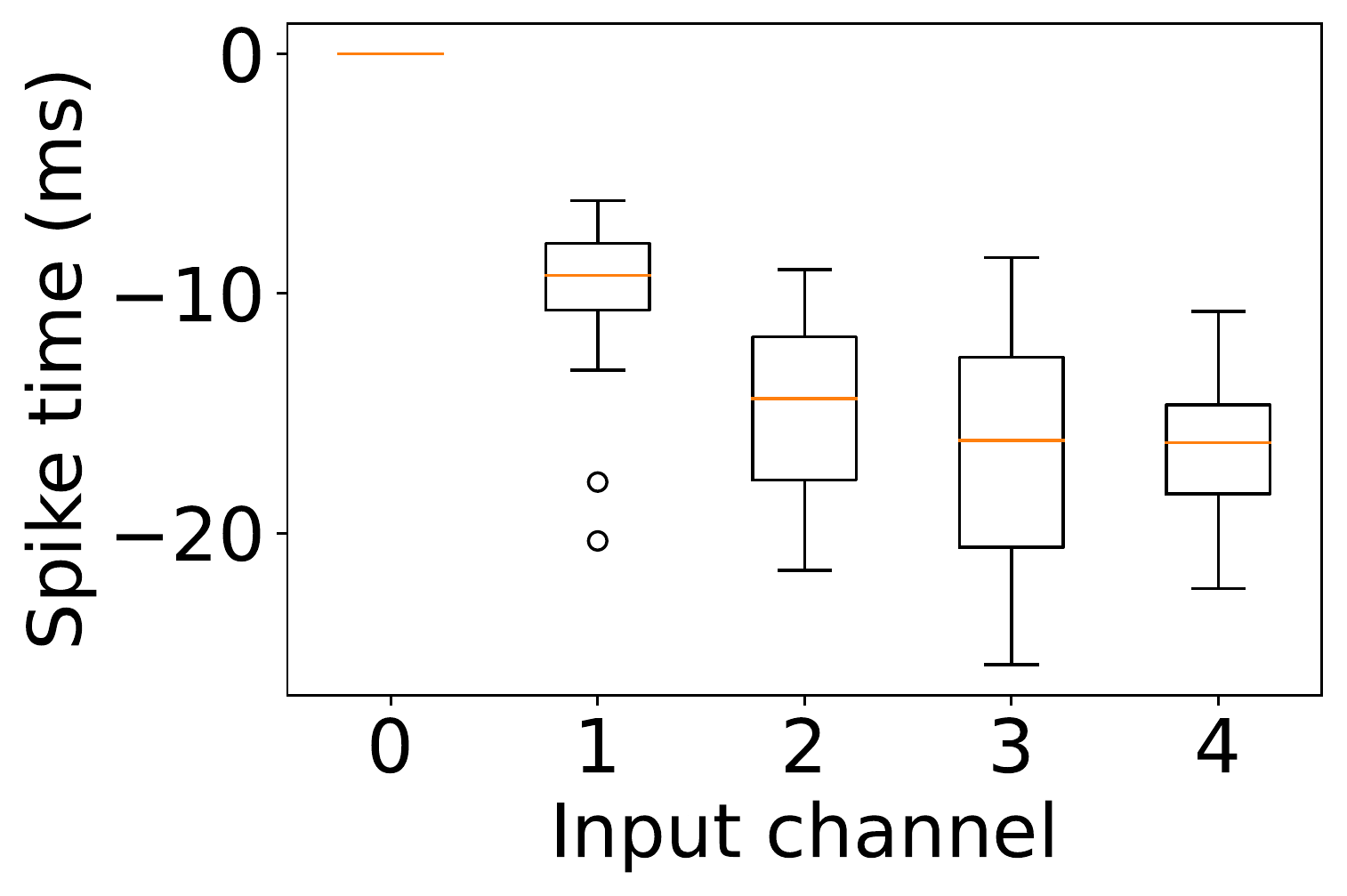}
        %\put(-95, 85){\textbf{B}}
        \caption{}
        \label{fig:rf_2}
    \end{subfigure}
    \\
    \begin{subfigure}[b]{0.49\columnwidth}
        \includegraphics[width=\textwidth]{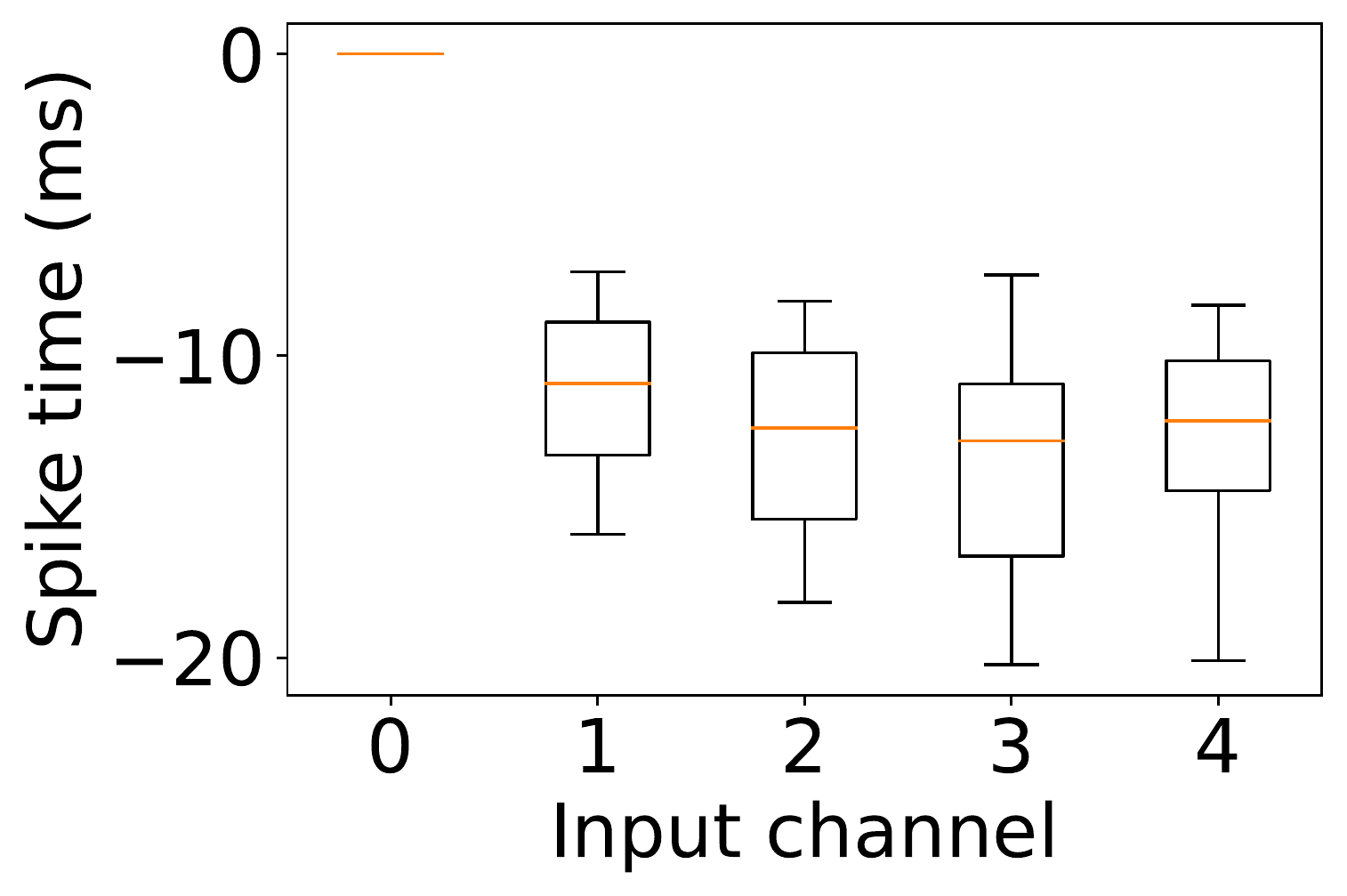}
        %\put(-95, 85){\textbf{C}}
        \caption{}
        \label{fig:rf_3}
    \end{subfigure}
    %\hfill
    \begin{subfigure}[b]{0.49\columnwidth}
        \includegraphics[width=\textwidth]{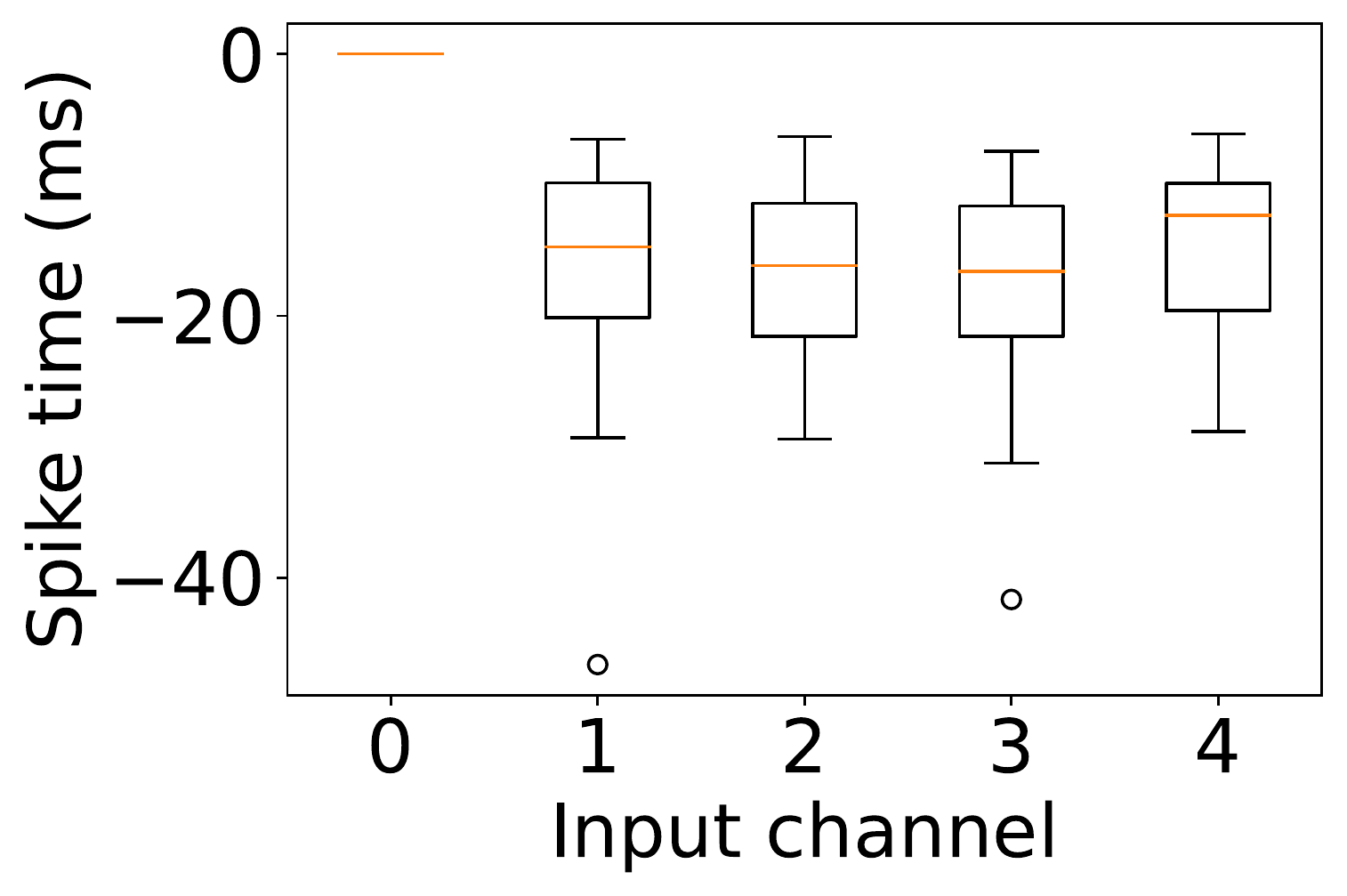}
        %\put(-95, 85){\textbf{D}}
        \caption{}
        \label{fig:rf_4}
    \end{subfigure}
    \caption{%
        \textbf{%
            Spatiotemporal receptive fields of four different B2 neurons.
        }%
        %The receptive fields were mapped by sampling with random stimuli (N~=~10,000).
        Each B2 neuron has a feed-forward input (Channel~0) at reference-time $t$~=~0 and four lateral inputs (Channels~1--4), each of which are stimulated with a single spike at a random time.
        %The receptive field of a neuron is estimated as the set of presynaptic spike patterns for which a postsynaptic spike is generated.
        Boxes extend from lower to upper quartiles, with lines at median presynaptic spike-times.
        Whiskers and flier points denote the range of data.
        Flier points are those past the end of the whiskers.
    }
    \label{fig:RFs}
\end{figure}

In this section, results are presented from the mapping of receptive fields of B2 neurons implemented in a DYNAP-SE processor, as described in Section~\ref{sec:RF_mapping}.
\textbf{Figure~\ref{fig:RFs}} illustrates the receptive fields of four different B2 neurons, in the form of box-and-whisker plots, aggregated from presynaptic stimulation with different randomized input patterns (N~=~10,000).
The plots consist of all the stimulus patterns---that is, channel--spike-time combinations, see \textbf{Figure~\ref{fig:rf_spikes}}---that made the B2 neuron in question generate at least one postsynaptic spike in response.
\textbf{Table~\ref{tab:RF_IDs}} presents the hardware neuron IDs---local to the used DYNAP-SE core---for which the receptive fields in \textbf{Figure~\ref{fig:RFs}} were observed.

\begin{table}[tb]
    \caption{%
        \textbf{%
            DYNAP-SE circuit-IDs used to obtain the different receptive fields.
        }%
        The neuron IDs are local to the processor core, and the synapse IDs are local to each neuron.
    }
    \centering
    \begin{tabular}{l c l}
        \toprule
        \textbf{Receptive field}    & \textbf{Neuron ID}    & \textbf{Synapse IDs}  \\
                                    & $\in$ [0,255]         & $\in$ [0,63]      \\
        \midrule
        Figure~\ref{fig:rf_1}       & 74    & 0--8 \\
        Figure~\ref{fig:rf_2}       & 129   & 0--8 \\
        Figure~\ref{fig:rf_3}       & 222   & 0--8 \\
        Figure~\ref{fig:rf_4}       & 248   & 0--8 \\
        \bottomrule
    \end{tabular}
    \label{tab:RF_IDs}
\end{table}

\subsection{Feature Tuning}
\label{sec:results:tuning}

\begin{figure}
    \centering
    \begin{subfigure}[b]{0.49\columnwidth}
        \includegraphics[width=\textwidth]{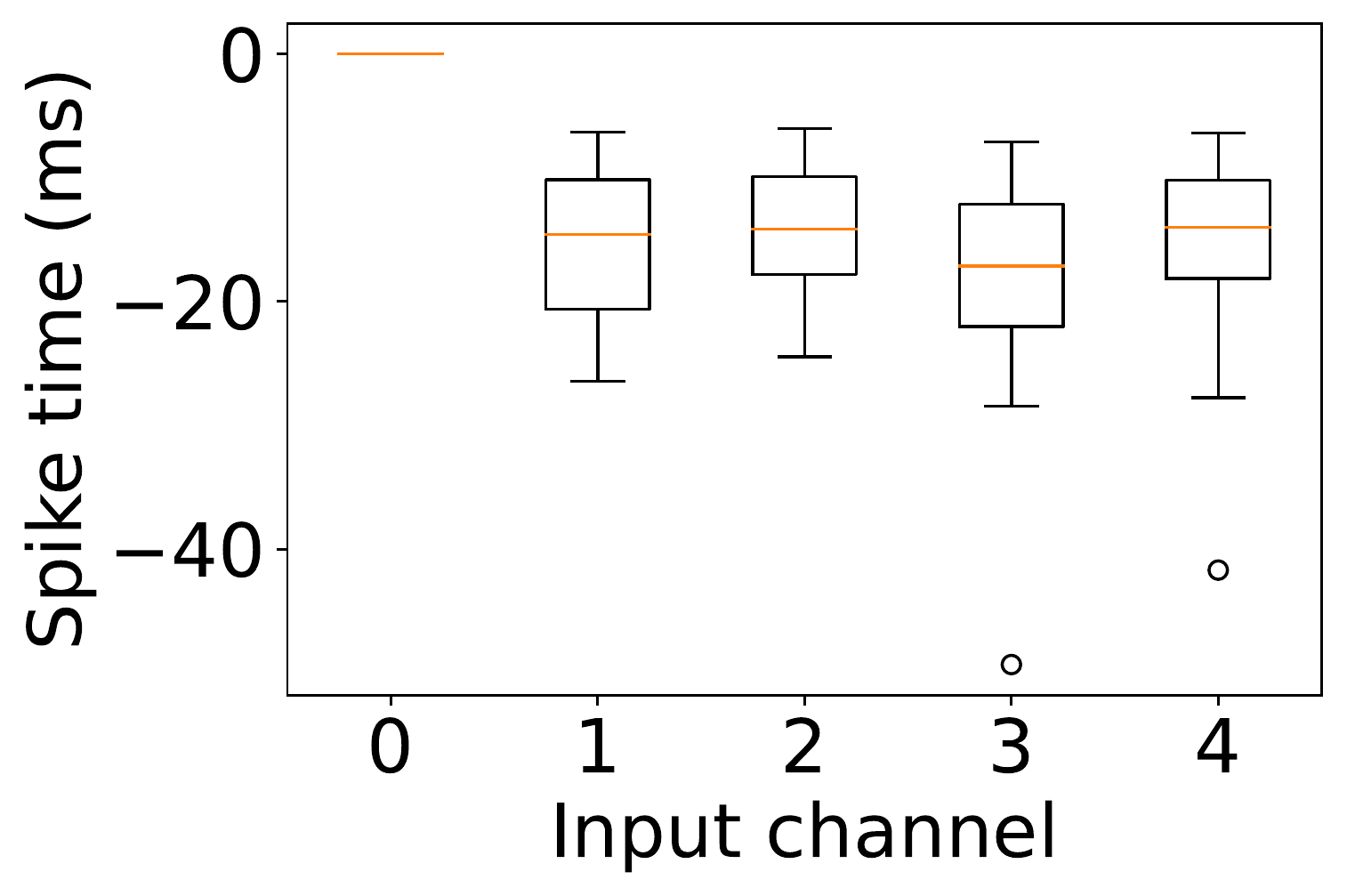}
        %\put(-95, 85){\textbf{C}}
        \caption{}
        \label{fig:rf_tuned_1}
    \end{subfigure}
    \begin{subfigure}[b]{0.49\columnwidth}
        \includegraphics[width=\textwidth]{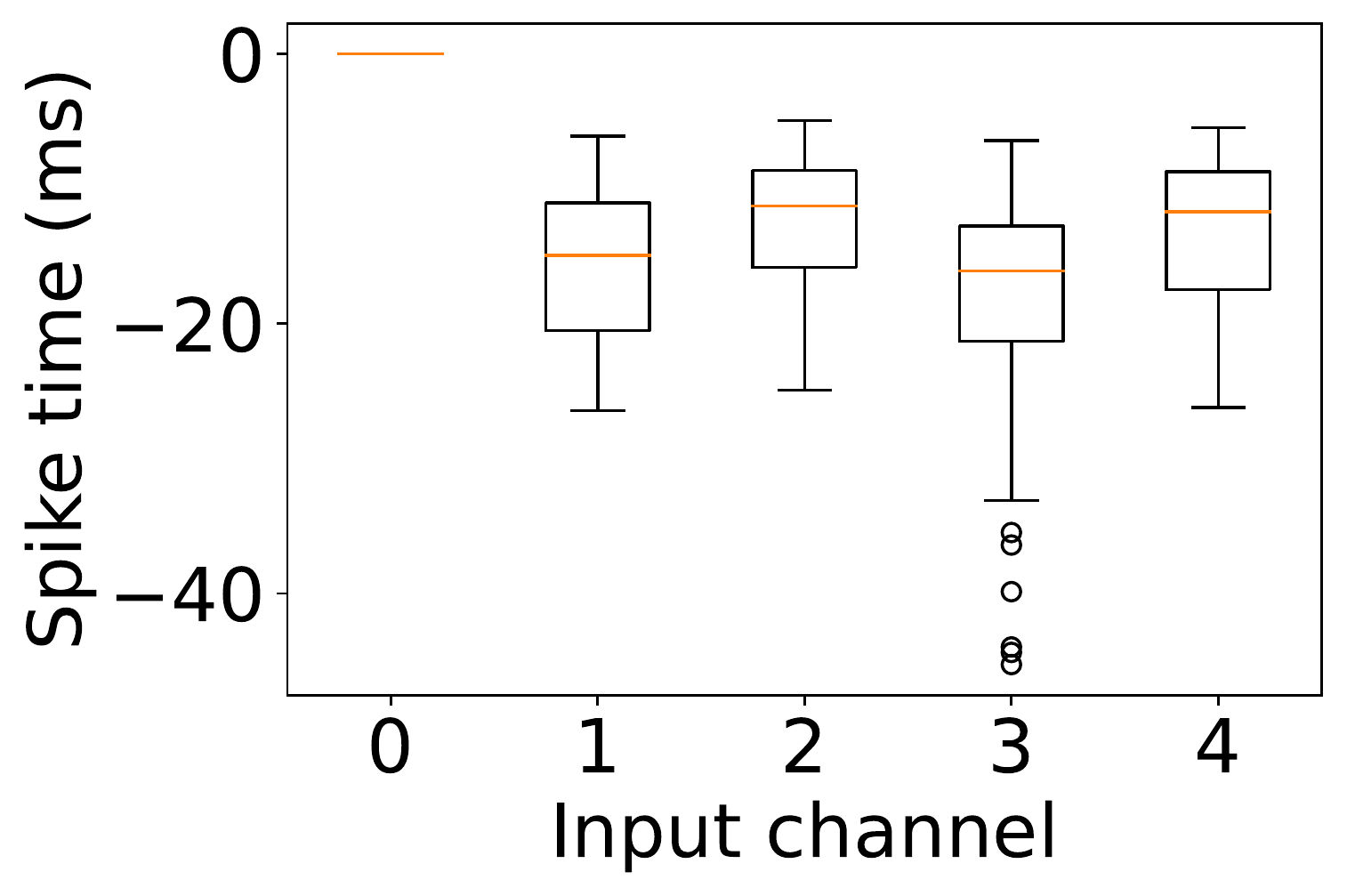}
        %\put(-95, 85){\textbf{D}}
        \caption{}
        \label{fig:rf_tuned_2}
    \end{subfigure}
    \caption{%
        \textbf{%
            Receptive fields resulting from feature tuning.
        }%
        The receptive field of the neuron prior to tuning is presented in Figure~\ref{fig:rf_4}.
    }
    \label{fig:rfs_tuned}
\end{figure}

\begin{figure}
    \centering
    \includegraphics[width=\columnwidth]{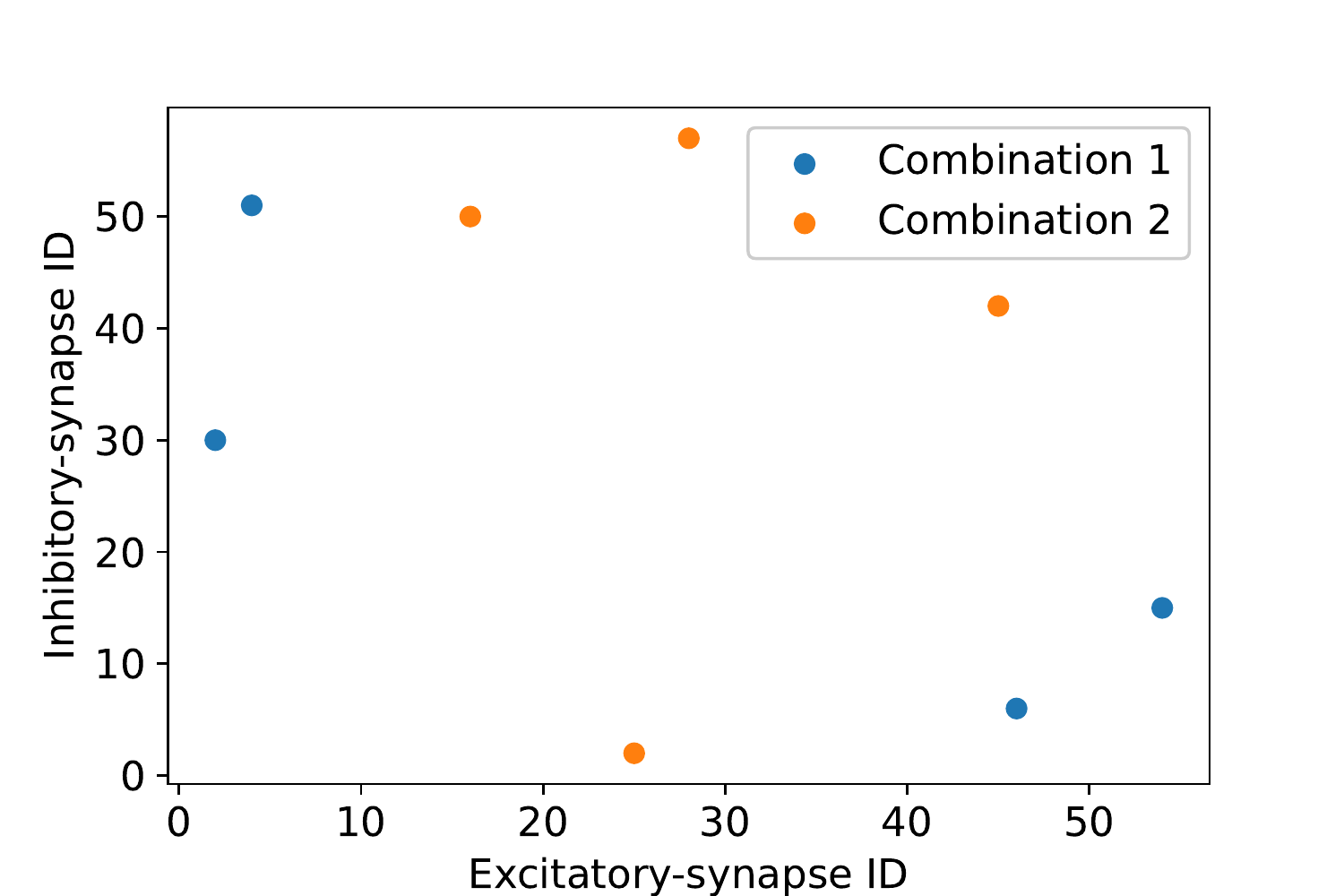}
    \caption{%
        \textbf{%
            Synapse-circuit combinations satisfying the feature tuning of a B2 neuron.
        }%
        Each dot represents the hardware synapse IDs $\in$ [0,63] used for the excitatory and inhibitory synapses of one of the four disynaptic elements, respectively.
        These synapse combinations correspond to the receptive fields presented in \textbf{Figure~\ref{fig:rfs_tuned}}.
    }
    \label{fig:cam_comb}
\end{figure}

The B2 neuron with the receptive field of \textbf{Figure~\ref{fig:rf_4}} was successfully reconfigured by synapse-circuit sampling to distinguish between Pattern~A and Pattern~B in \textbf{Figure~\ref{fig:tuning_patterns}} with a spiking and non-spiking response, respectively.
Prior to reconfiguration, the receptive field was fairly uniform, with the inputs on Channels 1--4 being almost interchangeable.
The feature tuning was accomplished by randomly replacing the synapse circuits used for the connections, see \textbf{Table~\ref{tab:tuned_IDs}} and \textbf{Figure~\ref{fig:cam_comb}}.
Out of 200 randomized configurations, four resulted in accurate discrimination between the two patterns.
\textbf{Figure~\ref{fig:rfs_tuned}} shows the receptive fields of two of these successful configurations.

\begin{table}[tb]
    \caption{%
        \textbf{%
            DYNAP-SE circuit-IDs resulting from feature tuning.
        }%
        The neuron IDs are local to the processor core, and the synapse IDs are local to each neuron.
        The synapse IDs used for each \gls{e-i} disynaptic connection are grouped inside parentheses.
        The original configuration of the receptive field of Figure~\ref{fig:rf_4} is included for reference.
    }
    \centering
    \begin{tabular}{l c l}
        \toprule
        \textbf{Receptive field}    & \textbf{Neuron ID}    & \textbf{Synapse IDs}  \\
                                    & $\in$ [0,255]         & $\in$ [0,63]      \\
        \midrule
        Figure~\ref{fig:rf_4}       & 248   & 0--8 \\
        \midrule
        Figure~\ref{fig:rf_tuned_1} & 248   & 56, (2, 30), (54, 15), (46, 6), (4, 51)   \\
        Figure~\ref{fig:rf_tuned_2} & 248   & 24, (25, 2), (45, 42), (28, 57), (16, 50) \\
        \bottomrule
    \end{tabular}
    \label{tab:tuned_IDs}
\end{table}

\subsection{Energy Usage}

\begin{figure}[tb]
    \centering
    \includegraphics[width=\columnwidth]{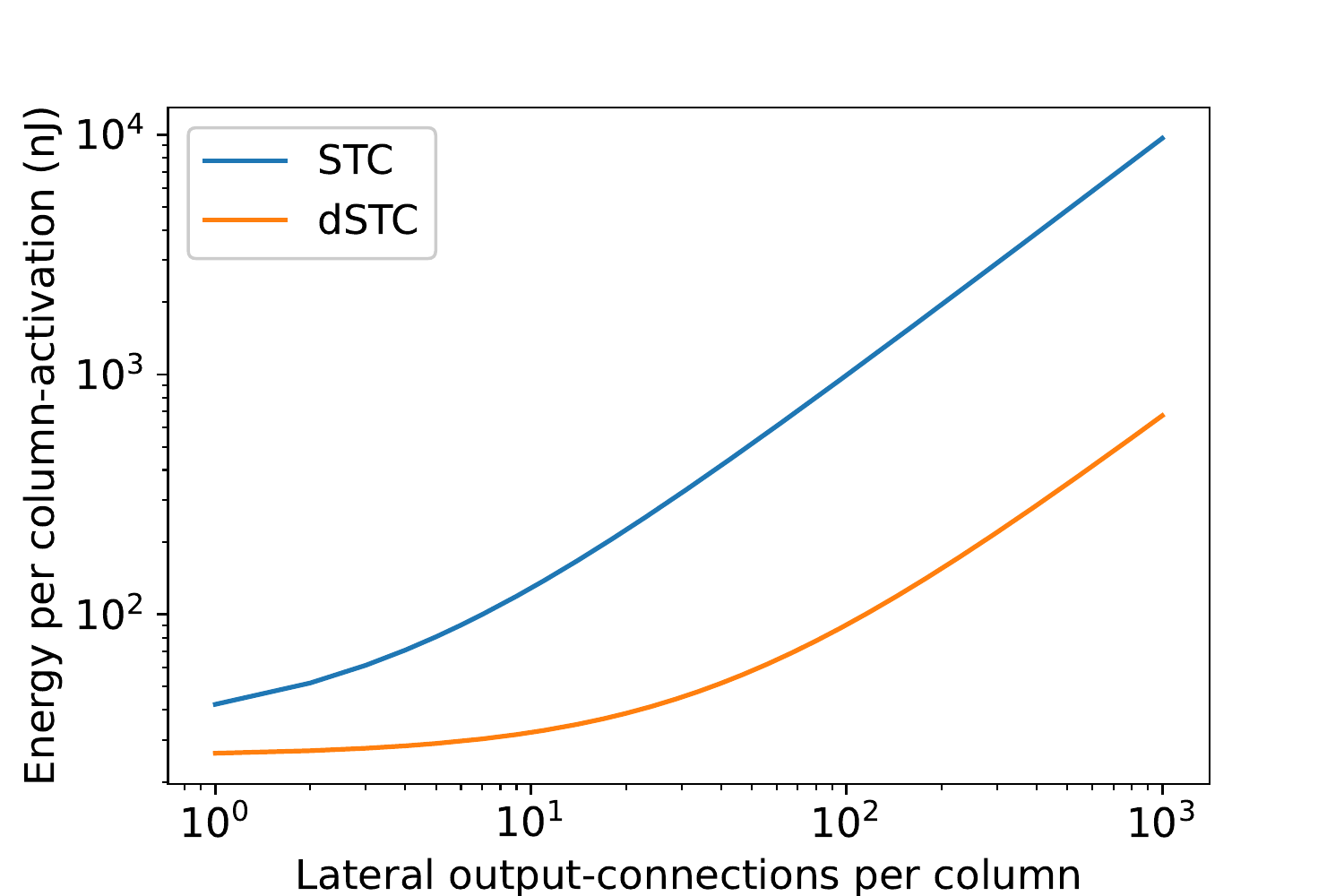}
    \caption{%
        \textbf{%
            STC network energy-scaling with number of lateral connections.
        }%
        The energy is estimated in terms of the spiking activity that results from an STC column-activation---that is, for each spike of an A neuron and the subsequently elicited intercolumnar spike-events.
    }
    \label{fig:energy}
\end{figure}

Based on measurements of the energy dissipation for the main operations of the DYNAP-SE \cite{moradi2018dynaps}, we present an estimate of the energy usage per laterally projected spike-event in \textbf{Table~\ref{tab:energy}} for the \gls{stc} and \gls{dstc}, respectively.
This estimate is made for the energy usage that is \textit{added} by each lateral spike to the cost of the already present spike generation and routing in the networks---which is motivated by the fact that the lateral connections constitute more than half of connections in the network and have the potential of scaling to even higher numbers.
The estimate suggests an improvement by one order of magnitude per lateral spike for the \gls{dstc} model---from 9.6 nJ to 0.65 nJ.
In \textbf{Figure~\ref{fig:energy}}, this energy comparison is put into the context of the energy usage of each whole network column and the energy-scaling with the number of lateral connections.

\begin{table}[tb]
    \caption{%
        \textbf{%
            Estimated energy usage per lateral spike.
        }%
        The energy measurements for the DYNAP-SE were retrieved from \cite{moradi2018dynaps}.
    }
    \centering
    \begin{tabular}{l l c c}
        \toprule
        \textbf{SNN}    & \textbf{Hardware}     & \textbf{Count}    & \textbf{Energy} \\
        \textbf{model}  & \textbf{operation}    &                   & \textbf{dissipation} \\
        \midrule
        STC     & Spike generation                  & 1     & 883 pJ    \\
                & Spike-and-destination encoding    & 1     & 883 pJ    \\
                & Intracore event-broadcast         & 1     & 6.84 nJ   \\
                & Intercore event-routing           & 1     & 360 pJ    \\
                & CAM-match pulse-extension         & 2     & 324 pJ    \\
        \cmidrule(r){2-4}
                & \multicolumn{2}{c}{\textbf{Sum}}  & \textbf{9.6 nJ}   \\
        \midrule
        dSTC    & CAM-match pulse-extension         & 2     & 324 pJ    \\
        \cmidrule(r){2-4}
                & \multicolumn{2}{c}{\textbf{Sum}}  & \textbf{0.65 nJ}  \\
        \bottomrule
    \end{tabular}
    \label{tab:energy}
\end{table}

\section{Discussion}
\label{sec:discussion}
% This should discuss the significance of the results and compare them with previous work using relevant references.

% Comparison to STC
In the \gls{dstc}---in contrast to former implementations of \gls{stc} networks---no interneuronal or axonal delays are required, since the dynamics of the postsynaptic \gls{e-i} currents contribute, in effect, delayed excitations that are unique and tunable for each lateral input connection.
Comparison with an \gls{stc} network having dedicated delay interneurons, as in \cite{sheik2012emergent}, shows a reduction of energy usage per lateral connection by about one order of magnitude, see \textbf{Table~\ref{tab:energy}}, for implementation on the DYNAP-SE.
\textbf{Figure~\ref{fig:energy}} illustrates how this difference is substantial already for the number of lateral connections investigated so far, and increasingly so for potentially larger numbers of such connections. 
Furthermore, the original implementation of the \gls{stc} used \textit{three} synapses for each lateral, delayed B2-input---in order to mitigate device mismatch with hardware redundancy---but the energy estimate presented here is conservatively made for the ideal case of one synapse per \gls{stc} connection.

% Evaluation of results/performance
The improved energy efficiency of the \gls{dstc} does, however, come at the cost of the relatively broadly tuned receptive fields of the output neurons, see \textbf{Figure~\ref{fig:RFs}}.
This is a trade-off that could be motivated, for example, in a deep neural network of stacked \gls{stc} layers with high fan-in on the B2 neurons.
Moreover, the temporal widths of the receptive fields are comparable to that of a coincidence-detection based feature-detection circuit found in crickets \cite{schoneich2015auditory}, which originally inspired the disynaptic elements \cite{sandin2020delays} used in the present work.
During the experiments, we observed some irregularities in the results, such as widening or narrowing of the receptive fields.
This variability is likely due to temperature effects \cite{song2021reliability} and may, for instance, be addressed with novel nanomaterials in future generations of neuromorphic hardware \cite{christensen2022roadmap}.

% Feature tuning
The feature-tuning experiment described in Section~\ref{sec:methods:tuning} provides proof-of-concept results, see Section~\ref{sec:results:tuning}, for hardware-circuit sampling as a possible approach for training or optimizing networks such as the \gls{dstc}.
Judging by the illustration of the resulting circuit combinations in \textbf{Figure~\ref{fig:cam_comb}}, these are fairly dissimilar to each other and span a large range of the possible circuit IDs.
This suggests that the solutions to the tuning problem investigated here are not unique, and, therefore, that the approach has potential for more complex tuning problems with a more constrained set of possible solutions.

\section{Conclusion}
\label{sec:conclusion}
% This section should be used to highlight the novelty and significance of the work, and any plans for future relevant work.

We have investigated spike-timing-based spatiotemporal receptive fields of single mixed-signal spiking neurons using heterogeneous synaptic dynamics \cite{bartolozzi2007synaptic} in the DYNAP-SE neuromorphic processor and the possibility of tuning such receptive fields by discrete synapse-address reprogramming.
The neurons were configured with balanced \gls{e-i} synaptic dynamics \cite{sandin2020delays, nilsson2020integration} and four lateral connections per neuron, such as the neurons in the output layer of an \gls{stc} network \cite{sheik2012emergent, sheik2013thesis}.
%
%We present an estimate of how such use of synaptic dynamics in the \gls{stc}, in place of its dedicated delay-neurons, lowers its energy-cost per lateral, delayed spike by one order of magnitude---from 9.6~nJ to 0.65~nJ.
%
We find that the energy-cost per spike on the lateral connections is reduced by about one order of magnitude---from 9.6~nJ to 0.65~nJ---when disynaptic dynamics are used instead of dedicated delay-neurons.
Furthermore, our conceptualization of the \gls{stc} in terms of the receptive fields of single neurons contributes a more detailed view of its pattern-recognition mechanism than the population-level analysis of previous work \cite{sheik2012emergent, coath2014robust}.
This may open up for further development of the network architecture, for instance by recognizing that the \gls{stc}'s receptive fields are similar to features in conventional deep neural networks.

%In summary, we have demonstrated how heterogeneous synaptic dynamics in mixed-signal neuromorphic processors, like the DYNAP-SE, can offer efficient mechanisms for event-driven, spike-timing-based spatiotemporal pattern recognition as a complement to more resource-intensive, delay-based coincidence-detection networks \cite{sheik2012emergent}, and biologically more plausible but resource-intensive models of dendritic integration \cite{wang2012active, kaiser2021emulating, benjamin2021neurogrid}.
%
%In future work, the influence on the receptive fields from maximization of the device-mismatch effects by further bias tuning could be investigated.
%
%Also, the effect of the inhibitory part of the disynaptic elements \cite{sandin2020delays} on the pattern selectivity could be more closely investigated, in contrast to the use of delayed pure excitations as in the original \gls{stc}.

\begin{acks}
This work was funded by
\grantsponsor{kempe}{The Kempe Foundations}{https://www.kempe.com/}
under contract~\grantnum{kempe}{JCK-1809}
and by
\grantsponsor{ecsel}{ECSEL JU}{https://www.kdt-ju.europa.eu/}
under grant no.~\grantnum{ecsel}{737 459}.
We thank Prof.\ Elisabetta Chicca for helpful feedback on an earlier version of this manuscript.
\end{acks}

\bibliographystyle{ACM-Reference-Format}
\bibliography{references}

\end{document}